\documentclass[twoside,11pt]{article}

\usepackage{blindtext}

\usepackage[[abbrvbib]{jmlr2e}
\usepackage{hyperref}
\usepackage{url}
\usepackage{graphicx}
\usepackage{booktabs}
\usepackage{amsmath}
\usepackage{pgfplots}
\usepackage{tikz}
\usepackage{caption}
\usepackage{multirow}
\usepackage{amsfonts}
\usepackage{bbold}
\usepackage{array}
\usepackage{soul}
\pgfplotsset{compat=1.13}% <- needed!!
\hypersetup{
    colorlinks,
    citecolor=black,
    filecolor=black,
    linkcolor=black,
    urlcolor=black
}

\usepackage{microtype}
\usepackage{subfigure}

\newcolumntype{P}[1]{>{\centering\arraybackslash}p{#1}}
\newcommand{\tablefontsize}[0]{
\fontsize{8pt}{10pt}\selectfont
}

\usepackage{lastpage}
\jmlrheading{24}{2023}{1-\pageref{LastPage}}{4/21; Revised
3/23}{5/23}{21-0445}{Xinchi Qiu, Titouan Parcollet, Javier Fernandez-Marques, Pedro P.\ B.\ Gusmao, Yan Gao, Daniel J. Beutel, Taner Topal, Akhil Mathur, Nicholas D. Lane}

\ShortHeadings{A First Look into the Carbon Footprint of Federated Learning}{Qiu,Parcollet,Fernandez-Marques,Gusmao,Gao,Beutal,Topal,Mathur,Lane}
\firstpageno{1}

\begin{document}

\title{A First Look into the Carbon Footprint \\ of Federated Learning}

\author{\name Xinchi Qiu$^{\star}$ \email  xq227@cam.ac.uk\\
        \name Titouan Parcollet$^{\dagger,\star}$ \email titouan.parcollet@univ-avignon.fr\\
        \name Javier Fernandez-Marques$^{\ddagger}$ \email javier.fernandezmarques@cs.ox.ac.uk\\
        \name Pedro P.\ B.\ Gusmao$^{\star}$ \email pp524@cam.ac.uk\\
        \name Yan Gao$^{\star}$ \email  yg381@cam.ac.uk\\
        \name Daniel J. Beutel$^{\mathsection,\star}$ \email db849@cam.ac.uk\\
        \name Taner Topal$^{\mathsection,\star}$ \email taner@flower.dev\\
        \name Akhil Mathur$^{\triangledown}$ \email akhil.mathur@nokia-bell-labs.com\\
        \name Nicholas D.\ Lane$^{\star,\mathsection}$ \email ndl32@cam.ac.uk
        \AND
        \addr $^{\star}$Department of Computer Science and Technology, University of Cambridge\\
        15 JJ Thomson Ave, Cambridge CB3 0FD, United Kingdom
        \AND 
        \addr $^{\dagger}$Laboratoire Informatique d'Avignon, Avignon Université\\
        339 Chemin des Meinajaries, 84000 Avignon, France
        \AND
        \addr $^{\ddagger}$Department of Computer Science, University of Oxford\\
        15 Parks Rd, Oxford OX1 3QD, United Kingdom
        \AND
        \addr $^{\mathsection}$Flower Labs GmbH\\
        Winterhuder Weg 29, 7. Stock, 22085 Hamburg, Germany
        \AND
        \addr $^{\triangledown}$Nokia Bell Labs\\
        21 JJ Thomson Avenue, Cambridge CB3 0FA, United Kingdom
       }
     
\editor{Qiaozhu Mei}

\maketitle

\begin{abstract}%   <- trailing '%' for backward compatibility of JMLR .sty file

Despite impressive results, deep learning-based technologies also raise severe privacy and environmental concerns induced by the training procedure often conducted in data centers. In response, alternatives to centralized training such as Federated Learning (FL) have emerged. FL is now starting to be deployed at a global scale by companies that must adhere to new legal demands and policies originating from governments and social groups advocating for privacy protection. \textit{However, the potential environmental impact related to FL remains unclear and unexplored. This article offers the first-ever systematic study of the carbon footprint of FL.} We propose a rigorous model to quantify the carbon footprint, hence facilitating the investigation of the relationship between FL design and carbon emissions. We also compare the carbon footprint of FL to traditional centralized learning. Our findings show that, depending on the configuration, FL can emit up to two order of magnitude more carbon than centralized training. However, in certain settings, it can be comparable to centralized learning due to the reduced energy consumption of embedded devices.
%despite being slower to converge in some cases, may result in a comparatively greener impact than a centralized equivalent setup. 
%We perform extensive experiments across different datasets, settings and various models with FL. 
Finally, we highlight and connect the results to the future challenges and trends in FL to reduce its environmental impact, including algorithms efficiency, hardware capabilities, and stronger industry transparency. 
%Recently, various companies started to deploy large-scale FL methods to meet with legal demands and policies originating from governments and the civil society for privacy protection, but also to benefit from a large amount of available distributed training data.
\\

\begin{keywords}
\vspace{+0.2cm}
federated learning, carbon footprint, energy analysis, green AI, on-device AI
\end{keywords}
\end{abstract}
\section{Introduction}
Atmospheric concentrations of carbon dioxide, methane, and nitrous oxide are at levels not seen in the last $800,000$ years \citep{adopted2014climate}. Together with other anthropogenic drivers, their effects have been detected throughout a network of distributed systems and are extremely likely to have been the dominant cause of the observed global warming since the mid-20$^{th}$ century \citep{pachauri2014synthesis,crowley2000causes}. Unfortunately, deep learning (DL) algorithms keep growing in complexity, and numerous ``state-of-the-art" models continue to emerge, each requiring a substantial amount of computational resources and energy, resulting in clear environmental costs \citep{nlp}. 
%Indeed, these models are trained for thousands of hours on specialized hardware accelerators such as Graphical Processing Units (GPU) in data centers that are extremely energy-consuming \citep{7966398}. 
Indeed, these models are routinely trained for thousands of hours on specialized hardware accelerators in data centers that are extremely energy-consuming \citep{7966398}. As \citet{amodei2018ai} showed, the amount of computing used by the largest machine learning (ML) training has been exponentially increasing and grown by more than $300,000\times$ from 2012 to 2018, which is equivalent to a 3.4-months doubling period -- a rate that dwarfs the well-known Moore's 2-year doubling period. Even though the amount of energy per FLOPS has been exponentially decreasing over time, making the deep learning model more and more computationally efficient, the carbon footprint of ML models is still one of the big concerns in society.

%This is an observation that forces us to consider the carbon footprint of deep learning methods.

%which is equivalent to compute requirements doubling every 3.4-months -- a rate that dwarfs typical hardware growth expectations such as Moore's law under which compute doubles only every 2-years. 

The data centers that enable DL research and commercial operations are not often accompanied by visual signs of pollution. In a few isolated cases, they are even powered by environmentally friendly energy sources \citep{googledatacentre,AWS}. Still, they are responsible for an increasingly significant carbon footprint. Each year data centers use $200$ terawatt-hours (TWh), which is more than the national electricity consumption of some countries, representing $0.3 \%$ of global carbon emissions \citep{nature,andrae2015global}. In comparison, the entire information and communications technology ecosystem accounts for $2 \%$. To put this issue in a more human perspective, each person on average on the planet is responsible for $5$ tonnes of emitted CO$_2$-equivalents (CO$_2$e) per year \citep{nlp}, while training a large Natural Language Processing (NLP) transformer model with neural architecture search may produce $284$ tonnes of CO$_2$e \citep{nlp}. Even for smaller deep neural networks and routine research experiments, \cite{parcollet21_interspeech} demonstrated that the training process necessary to create a state-of-the-art speech recognizer could produce more than $0.1$ tonnes of CO$_2$e with consumer-grade hardware. Even though the number refers to one of the largest ML models, given the increasing interest in Large Language Models (LLM), it is likely that this trend will continue and possibly expand to tasks besides NLP. Understanding the carbon footprint of ML training will play a paramount role in allowing people to develop more carbon-efficient models and hardware, making the emission more transparent, and choosing renewable energy where possible.

%introducing FL
Decentralized alternatives to a data center-based DL and other forms of machine learning are emerging. Among these, the most prominent to date is \textit{Federated Learning (FL)}, first formalized by \citet{pmlr-v54-mcmahan17a}. %\cite{mcmahan2017federated}.
Under FL, training of models primarily occurs in a distributed scenario, either across a large number of personal devices (\emph{cross-device}), such as smartphones; or across a small number of institutions that cannot share data among themselves (\emph{cross-silo}), such as private hospitals. Devices collaboratively learn a global model but do so without uploading to a data center any of the locally stored sensitive data. Then they send the locally trained models to a central server, where models get aggregated following a strategy such as FedAVG \citep{mcmahan2017federated,kairouz2019advances,konevcny2015federated}. While FL is still a maturing technology, it is already being used by millions of users on a daily basis; for example, Google uses FL to train models for: predictive keyboard, device setting recommendation, and hot keyword personalization on phones \citep{mcmahan2017federated}. 

% why FL
At present, data owners are holding more and more sensitive information, such as individual activity data, life-logging videos, email conversations, and others \citep{fedcs}, so keeping personal medical and healthcare data private recently became one of the major ethical concerns \citep{kish2015unpatients}. To this extent, and in response to an increasing number of such privacy issues, policy-makers have responded with the implementation of data privacy legislation such as the European General Data Protection Regulation (GDPR) \citep{lim2020federated}. Due to these regulations, moving data across national borders becomes subject to data sovereignty law, making centralized training infeasible in some scenarios \citep{hsieh2020non}.

Furthermore, there are nearly seven billion connected Internet of Things (IoT) devices \citep{lim2020federated} and three billion smartphones around the world, potentially giving access to an astonishing amount of training data and decentralized computing power for meaningful research and applications. sing mobile sensing and smartphones to boost large-scale health studies, such as in \citet{7167448}, \citet{9106648} and \citet{shen2015smartphones}, has caused increased interest in the healthcare research field, and privacy-friendly framework including FL are potential solutions to answer this demand.

%These IoT devices normally consume an order of magnitude less power than those typically used in data centers. Using mobile sensing and smartphones to boost large-scale health studies, such as in \citet{7167448}, \citet{9106648} and \citet{shen2015smartphones}, has caused increased interest in the healthcare research field, and privacy-friendly framework including FL are potential solutions to answer this demand.% Also, IoT devices are usually built using devices that consumed much less power than GPUs that often are deployed in datacenters.

Despite FL privacy being under great scrutiny from the scientific community, we currently have little to no understanding of its impact on carbon emissions. This is a worrying situation, given the increasing interest in this technology. Therefore, the carbon footprint of FL needs to be assessed before vast systems are further deployed.
%We must not repeat the same mistakes made with the datacenters, whereby vast systems were deployed and became an everyday part of life, many years before the environmental consequences were assessed.

% talking about Carbon footprint
Whilst the carbon footprint for centralized learning has been studied in many previous works \citep{anthony2020carbontracker,lacoste2019quantifying,henderson2020towards,uchechukwu2014energy}, the energy consumption and carbon footprint related to FL remains virtually unexplored. This article provides the key step in attempting to fill this void by giving a first look into the carbon analysis of FL.
It expands upon our initial treatment of the area \citep{qiu2021federated} with a more comprehensive study; our original paper, and this article, have also prompted significant subsequent investigations within the community \citep{3480129, 9807354, 10032558}. 
%\hl{less} unexplored. This paper attempts to fill such void. %To this extent, this paper proposes to overcome this issue by giving a first look into the carbon analysis of FL.
Studies of this kind are essential because state-of-the-art results in deep learning are usually determined by metrics such as the accuracy of a given model or model size, while energy efficiency is often overlooked. Whilst accuracy remains crucial, we hope to encourage researchers to also focus on other metrics that are in line with the increasing societal global warming awareness. Recent research \citep{patterson2022carbon} indicates the approaches to reduce energy and carbon emissions in centralized training in data centers. By quantifying carbon emissions for FL and demonstrating that very specific FL setups may lead to a decrease of these emissions, we encourage the integration of the released CO$_2$e as a crucial metric to the FL deployment.  
%This paper addresses this looming gap in the literature. Whereas our still nascent understanding of the environmental impact of machine learning is isolated to datacenters -- we advance this situation by considering specifically FL. Existing results \citep{lacoste2019quantifying, anthony2020carbontracker} propose approaches that consider CO$_2$e emissions only under centralized training assumptions. None of these tools generalize to FL which this work aims to provide: a methodology to properly estimate CO$_2$e emissions related to FL. 
The scientific contributions of this work are as follows: 
\begin{itemize}
    \item \textbf{Analytical Carbon Footprint Model for FL. } 
    We provide the first quantitative CO$_2$e emissions estimation method for FL (Section \ref{secdatacentre}), including emissions resulting from both hardware training and communication between server and clients. % The carbon footprint quantified in this paper addresses emissions resulting from the FL training process only. It does not include emissions related to hardware manufacturing as such information is still largely unavailable.
    
    \item \textbf{Extensive Experiments.} Carbon sensitivity analysis is conducted with this method on real FL hardware under different settings, strategies, and tasks (Section \ref{secexp}). We demonstrate that CO$_2$e emissions depend on a wide range of hyper-parameters and that emissions derived from communication between clients and server can represent from $0.7\%$ up to more than $96\%$ of total emission. When compared to centralized training, we show that for different tasks and settings, FL can emit from $72\%$ to hundreds of times more carbon than its centralized version. %We also show that by considering more efficient strategies, we can reduce CO$_2$e emissions up to \hl{$60\%$}. 
    \item 
    \textbf{Analysis and Roadmap towards Carbon-friendly FL. } 
    We provide a comprehensive analysis and discussion of the results to highlight the challenges and future research directions in developing carbon-friendly federated learning (Section \ref{secanalysis}). 
\end{itemize}

\section{Federated Learning Background}\label{secbackground}

% what is FL
Traditional machine learning involves using a central server that hosts the machine learning models and all the data in one place. In contrast, in FL frameworks client devices collaboratively learn a shared global model using their own local data. FL has distinct privacy advantages over centralized training as the data are not transferred to the central server for training. In fact, the only information transferred from clients to the server is their respective updated model parameters obtained after each local training. To further limit the leakage of client's information in the model update, several mechanisms have been proposed over the years including Secure Aggregation~\citep{45808} and Differential Privacy~\citep{brendan2018learning}.

FL training occurs over multiple communication rounds. During each round, a fraction of the clients are selected and receive the global model from the server. Those selected clients then perform local training with their local data before sending the updated models back to the central server. Finally, the central server aggregates these updated models, resulting in a new global model. Then, this three-stage process is repeated for a fixed number of rounds.

There exists several aggregation strategies targeting to solve different FL problems. The most widely adopted one is FedAvg \citep{pmlr-v54-mcmahan17a}, in which the central server aggregates the models by performing a weighted sum of the received parameters based on the number of samples in each local dataset. More advanced strategies inspired by adaptive momentum-based gradient descent optimizers have also been proposed e.g., FedADAM \citep{reddi2020adaptive}.

%In addition, carbon footprint of running a machine learning model will mainly depend on the particular hardwares that the model is deployed on. We describe the specific hardwares that will be using in our experiments below. 

In addition, FL settings can be classified as either \emph{cross-silo} or \emph{cross-device}. In a \emph{cross-silo} scenario, clients are generally few, with high availability during all rounds, and are likely to have similar data distribution for training, e.g.\ consortium of hospitals. This scenario serves as motivation to consider Independent and Identically Distributed (IID) distributions. On the other hand, a \emph{cross-device} system will likely encompass thousands of clients having very different data distributions (non-IID) participating in just a few rounds, e.g.\ training of next-word prediction models on mobile devices. In practice, non-IID datasets not only means class imbalance, but also feature imbalanced among clients. Indeed, many latent factors can change such as the voice timbre in speech recognition \citep{gao2022end}.

%\textbf{FL strategy.} To better reflect realistic FL scenarios, we propose to investigate the energy consumption with the common FedAVG strategy \citep{pmlr-v54-mcmahan17a}, and more complex FedADAM strategy \citep{reddi2020adaptive}. For CIFAR10, we follow the experimental protocol proposed in \citet{reddi2020adaptive} considering the suggested best values for $\eta$, $\eta_l$, and $\tau$ in almost every experiment except for FedAVG, where we had to lower the value of $\eta_l$ to $10^{-3/2}$ to allow training. All other experiments used a server learning rate $\eta = 0.1$ and $\tau= 0.001$.
%\input{2.2_motivation}
\section{Quantifying CO$_2$e emissions}
\label{secdatacentre}

%When quantifying the CO$_2$e emission associated with a given process, it is important to consider not only the CO$_2$e being produced, but also the CO$_2$e embodied in the infrastructure being used. On the other hand, it is also important to avoid multiple accounts of the same emission when shared resources are considered. In those cases, embodied CO$_2$ accountability should be shared based on usage.   

%With these in mind, we consider both the CO$_2$e due to training and server-client data communication. Given the opportunistic nature of cross-device FL, we refrain from accounting for the CO$_2$e embodied in mobile devices as training is typically not their predominant usage.

Two major steps can be followed to quantify the carbon footprint of training deep learning models either in data centers or on the edge. First, we perform an analysis of the energy required by the method (Section \ref{secbasicelements}), mostly accounting for the total amount of energy consumed by the hardware. It includes training energy for centralized learning and training and communication energy for FL (Section \ref{secwan}). Then, the latter amount is converted to CO$_2$e emissions (Section \ref{secconversion}) based on geographical locations which, as it will be presented, vary significantly depending on the sources of energy. This study does not include emissions related to hardware manufacturing as such information is still largely unavailable.

\subsection{Training Energy Consumption}\label{secbasicelements}
First, we consider the energy consumption coming from GPU and CPU, which can be measured by sampling GPU and CPU power consumption at training time \citep{nlp}. For NVIDIA-based hardware, we can repeatedly query the NVIDIA System Management Interface (NVIDIA-smi) to sample the GPU power consumption and report the average over all processed samples while training. In the context of FL, not all clients are equipped with a GPU, and this part can thus be removed from the equation if necessary. To this extent, we propose to consider $e_{clt}$ as the power of a single client combining both GPU and CPU measurements. Then, we can connect these measurements to the total training time of the model. We define $T_{FL}(e,N,R)$ to be the total training energy consumption consisting of a total of $N$ clients in the pool with hardware power $e$ for a total of $R$ rounds in FL setup:
%the energy consumed for training total of $n$ clients in each communication round with wall clock time per round $t_i$ and power $e_{client,i}$ of each client:

%In addition, federated learning is the training strategy that we can fully utilise the huge amount of data in the devices without making clients to transfer their data to the server, so we do not need to consider the energy required to transferring data to server. On the other hand, during each communication round, a global model is sent from server to each chosen clients to be trained, hence the energy required to transfer the model parameters need to be included in the estimation. Let $m$ GB be the size of model parameters, the energy needed for transferring the parameters will be $5m$ kWh.

\vspace{-0.5cm}
\begin{align}
\label{eqenergy_fl}
\operatorname{T}_{FL}(e,N,R) = \sum_{j=1}^{R} \sum_{i=1}^{N} \mathbb{1}_{\{ Clt_{i,j}
\}} \cdot t_i \cdot e_{client,i},
\end{align}
where $\mathbb{1}_{\{Clt_{i,j}\}}$ is the indicator function indicating if client $i$ is chosen for training at round $j$, $t_i$ the wall clock time per round and $e_{clt,i}$ the power of client $i$.

Hardware components, such as system memory and storage, are also responsible for energy consumption. According to \citet{hodak2019towards}, one may expect a variation of around $10\%$ while considering these parameters. However, they are also highly dependent on the infrastructure considered and the device distribution that is unfortunately unavailable. We exclude the energy costs of powering such components since they account for a small portion of the total energy consumption during training. \\ 

\noindent\textbf{The particular case of cooling in centralized training.}
Cooling in data centers accounts for up to $40\%$ of the total energy consumed \citep{capozzoli2015cooling}. While this parameter does not exist for FL, it is crucial to consider it when estimating the cost of centralized training. Such estimation is particularly challenging as it depends on the data center efficiency. To this extent, we consider the use of Power Usage Effectiveness (PUE) ratio. As reported in the \textit{2019 Data Center Industry Survey Results} \citep{UptimeInstitute}, the world average PUE for the year 2019 is $1.67$. As expected, observed PUE strongly varies depending on the considered company. For instance, \textit{Google} declares a comprehensive trailing twelve-month PUE ratio of $1.11$ \citep{Google} compared to $1.2$ and $1.125$ for \textit{Amazon} \citep{AWS} and \textit{Microsoft} \citep{Microsoft} respectively. We also report a PUE ratio of a University-scale cluster (Avignon University, France) as an example. The PUE ratio is reported to be $1.55$ for a cluster containing $17$ computing nodes with $4$ to $8$ GPUs each. Therefore, Eq. (\ref{eqenergy_fl}) is adapted to centralized training setting as:

\vspace{-0.25cm}
\begin{equation}
\label{eqenergy_ct}
\operatorname{T}_{center} = \operatorname{PUE} \cdot (t \cdot e_{center}),
\end{equation}

with $e_{center}$ representing the power combining both GPUs and CPUs in a centralized training setup, and $t$ stands for the total training time.

\subsection{Wide-area-networking (WAN) Emission}
\label{secwan}

As clients continue to perform individual training on local datasets, their models begin to diverge. To mitigate this effect, model aggregation must be performed by the server in a process that requires frequent exchange of models between clients and the server. 

%For this reason, we must also consider the amount of emission derived from data exchange between server and clients during the beginning and the end of a round.
According to \citet{malmodin2018energy}, the embodied carbon footprint for Information and Communication Technology (ICT) network operators is mainly related to the construction and deployment of the network infrastructure including digging down cable ducts and raising antenna towers. 

Regarding FL, we estimate the energy required to transferring model parameters between the server and the clients following two parts. The first part is the energy consumed by routers throughout the FL communication process, while the second part is the energy consumed by the hardware when downloading and uploading the model parameters. We propose to use country-specific download and upload speed as reported on \textit{Speedtest} \citep{Speedtest} and router power reported on \textit{The Power Consumption Database} \citep{router}. Due to the rapid development of ICTs, we propose to use the median power obtained from all data submitted during 2021 to the database. We also take idle power consumption of hardware into consideration while they are communicating model parameters. Let us define $D$ and $U$ the download and upload speeds expressed in Mbps respectively. The communication energy per round is defined as:
\begin{equation}
\label{eqnetwork}
\operatorname{C}(e,N,R) = 
\sum_{j=1}^{R} \sum_{i=1}^{N}\mathbb{1}_{\{ Clt_{i,j}\}} \cdot S \cdot \left(\frac{1}{D}+\frac{1}{U}\right)\cdot (e_{r}+e_{idle,i}),
\end{equation}

with $S$ the size of the model in Mb, $e_{r}$ the power of the router, and $e_{idle}$ the power of the hardware of the idle clients.

\subsection{Converting to CO$_2$e emissions}\label{secconversion}
\begin{figure}[!ht]
%\caption{Global heat map of electricity to CO$_2$e conversion rate (in kg/kWh). The conversion rate are obtained from governmental sources or on the website Climate Transparency \footnote{Climate Transparency: \url{https://www.climate-transparency.org/}}.}
\caption[Caption for LOF]{Global heat map of electricity to CO$_2$e conversion rate (in kg/kWh). The conversion rate are obtained from governmental sources or on the website Climate Transparency \protect\footnotemark}
\centering
\includegraphics[width=0.7\textwidth]{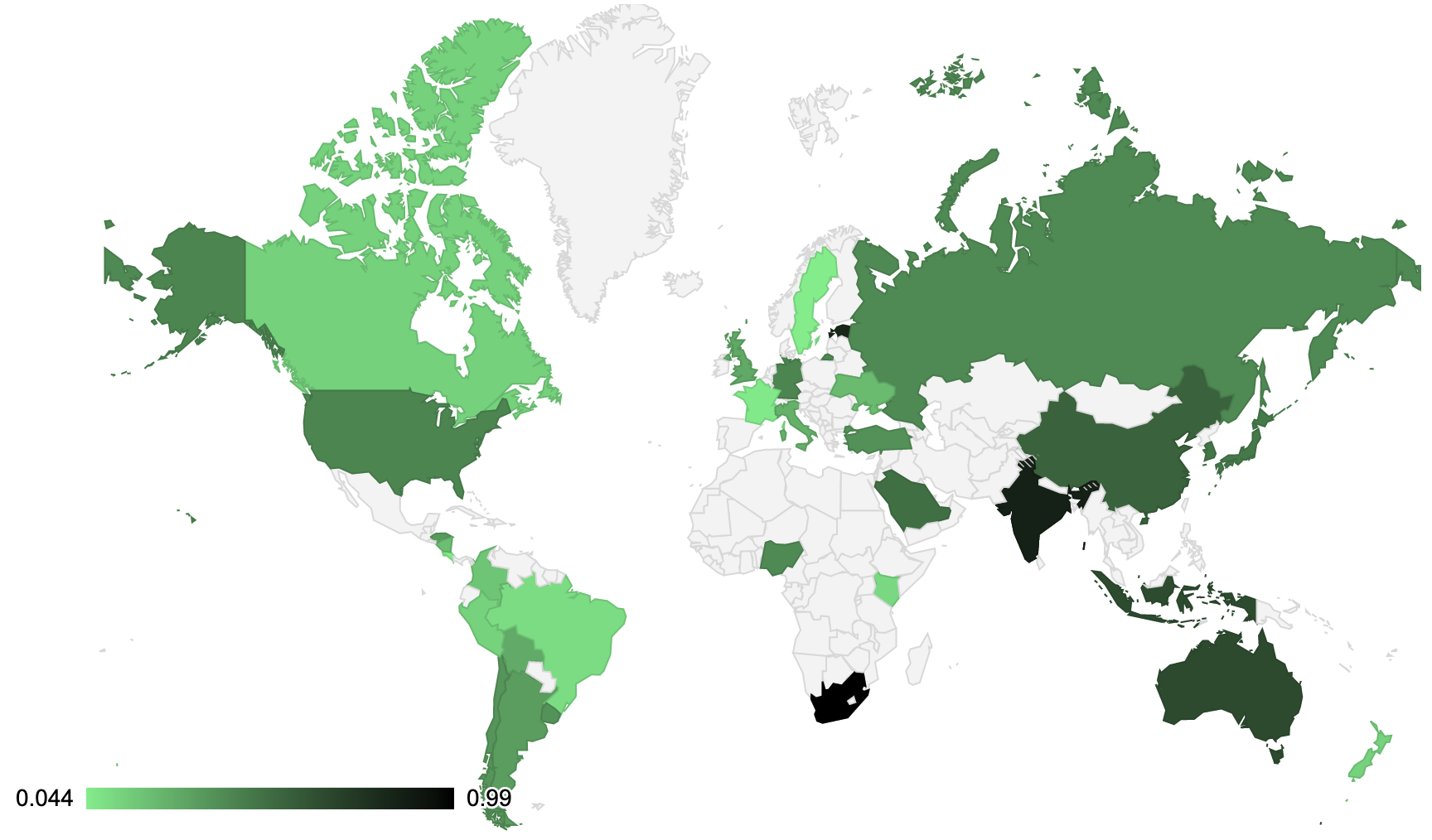}
\label{figheatmap}
\end{figure}

Realistically, it is challenging to compute the exact amount of CO$_2$e emitted in a given location since the information regarding the energy grid, \textit{i.e.}, the conversion rate from energy to CO$_2$e, is rarely publicly available \citep{lacoste2019quantifying,hodak2019towards}. Therefore, we assume that all data centers and the edge devices are connected to their local grid directly linked to their physical location. Electricity-specific CO$_2$e emission factors are obtained from official governmental websites and reports. Out of all these conversion factors expressed in \textit{kg CO$_2$e/kWh}, we picked three of the most representative ones averages over a one year-period: Australia ($0.656$)\footnote{source:\url{https://www.climate-transparency.org/countries/asia/australia}}, the United Kingdom ($0.281$) \footnote{source:\url{https://www.climate-transparency.org/countries/europe/the-united-kingdom}} and France ($0.054$)\footnote{source:\url{https://www.climate-transparency.org/countries/europe/france}}. The estimation methodology provided takes into accounts both transmission and distribution emission factors (\textit{i.e.} energy lost when transmitting and distributing electricity) and the efficiency of power plants. As expected, countries relying on carbon-efficient productions are able to lower their corresponding emission factor (\textit{e.g.} France, Canada). A heatmap demonstrating different levels of conversion rates in various countries can be found in Fig. \ref{figheatmap}.
%France among all countries has the lowest factor due to the fact that $75 \%$ of its electricity derives from nuclear power, which does not directly produces CO$_2$e. Other countries such as China, has one of the highest factor, due to a large part of coal-based energy production ($60 \%$). 

Therefore, the total amount of CO$_2$e emitted in kilograms for FL ($E_{FL}$) and centralized training  ($E_{center}$) obtained from Eq. \ref{eqenergy_fl}, \ref{eqenergy_ct} and \ref{eqnetwork} are:

\vspace{-0.25cm}
\begin{align}\label{eqcarboneq}
&\operatorname{E}_{FL}=  \operatorname{c}_{rate} \cdot [ \operatorname{T}(e,N,R) + \operatorname{C}(e,N,R) ],\\\label{eqcarboneq2}
&\operatorname{E}_{center}=  \operatorname{c}_{rate} \cdot \operatorname{T}_{center},
\end{align}
where $c_{rate}$ is the conversion rate factor. It is worth noticing that when dealing with non-IID partitions, the total training energy consumption ($\operatorname{T}(e,N,R)$) and energy for communication cost ($\operatorname{C}(e,N,R)$) will often be larger than IID partitions, as it usually requires larger number of communication rounds to reach certain model performance \cite{karimireddy2020scaffold,zhao2018federated}, and it is also shown in our experiments later in Section \ref{secexp}. In general, $c_{rate}$ will depend on the physical location of the hardware where the training takes place, and it is possible that $c_{rate}$ is not unique across the FL settings as clients can be scattered around the globe. We will need to adjust the $c_{rate}$ for each client based on their physical locations. In our experiments, we assume that all FL clients are located at the same physical locations for ease of comparison.

Carbon emissions may be compensated by carbon offsetting or with the purchases of Renewable Energy Credits (RECs, in the US) or Tradable Green Certificates (TGCs, in the EU). Carbon offsetting allows polluting actions to be mitigated directly via different investments in \textit{carbon-friendly} projects, such as renewable energies or massive tree planting \citep{anderson2012inconvenient}. RECs and TGCs \citep{bertoldi2006tradable}, on the other hand, guarantee that specifics volumes of electricity are generated from renewable energy sources. However, in our analysis, carbon rates are obtained at country level and do not integrate industry level carbon offsetting schemes or RECs. 
%More precisely, RECs and TGCs purchasing aims to create more renewable energy in the long-term according to the \textit{additionality} principle, by creating an increased demand for these energy sources. 
\footnotetext{Climate Transparency: \url{https://www.climate-transparency.org/}}

\section{Experiments}\label{secexp}

This article provides extensive estimates across different types of tasks and datasets, including image classification with CIFAR10 \citep{krizhevsky2009learning}, FEMNIST~\citep{mnist,cohen2017emnist}, and ImageNet \citep{russakovsky2015imagenet}, speech processing with keyword spotting on Speech Commands \citep{speechcommands} and speech recognition with CommonVoice \citep{ardila2020common}.  %We follow the hardware, data partition methodologies explained in Section \ref{sec:background}.
First, we provide an estimate of the carbon footprint following different realistic FL setups. Then, we conduct an in-depth analysis of these results to highlight the differences observed.

\subsection{Experimental Protocol} \label{secprotocol}
Experiments are built on top of PyTorch~\citep{NEURIPS2019_9015} and SpeechBrain \citep{ravanelli2021speechbrain}. We make use of the Flower framework~\citep{beutel2020flower} to implement and parameterized different FL training pipelines.
In addition to the carbon model (Section \ref{secdatacentre}), results are influenced by configurations of the hardware and systems of datacenters and FL respectively.%, which is explained in Section \ref{secbackground}. The FL pipeline is implemented using the Flower framework \citep{beutel2020flower}, SpeechBrain \citep{ravanelli2021speechbrain} and PyTorch~\citep{NEURIPS2019_9015}.

\textbf{Centralized training hardware.} We run our experiments on a server equipped with two Xeon 6152 22-core processors and NVIDIA Tesla V100 32GB GPUs. The CPU and GPU have TDP of 240W and 250W, respectively. We use a single GPU per experiment and measure the power drawn by both CPU and GPU through  \textit{nvidia-smi} monitoring and the cross-platform \textit{psutil} tools.

\textbf{Federated learning hardware.} We consider the use of NVIDIA Tegra X2 \citep{tegrax2} and Jetson Xavier NX \citep{nx} devices as our FL clients. These devices can be viewed as a realistic pool of FL clients since they can be found embedded in various IoT devices including cars, smartphones, and video game consoles. NVIDIA Tegra X2 offers two power modes with theoretical power limits of $7.5 W$ and $15W$ and Xavier NX offers $10 W$ and $15W$. Across our different runs, we use the lower power mode for each device, and we employ the built-in utility \textit{tegrastats} to report the overall power consumption.
For both power consumption and training time, we report averaged values across several FL rounds for each experiment. We also measure the idle power consumption for both devices, which was recorded as $1.35W$ and $2.25W$ for TX2 and NX respectively.

\textbf{Datasets.} We conduct our estimations on three image classification tasks of different complexity both in terms of the number of samples and number of classes: CIFAR10, FEMNIST, and ImageNet. FEMNIST, federated extended MNIST, is built by partitioning the data in Extended MNIST (EMNIST) according to writer ids. It contains 671K 28$\times$28 images of digits and letters. In addition, we also perform analysis on speech processing following the same complexity concern with the Speech Commands dataset for keyword spotting and Common Voice for automatic speech recognition (ASR). Speech Commands contains $65K$ 1-second long audio clips of $30$ keywords, with each clip consisting of only one keyword. Following the setup described in \citet{zhang2017hello}, we train the model to classify the audio clips into one of the $10$ keywords - ``Yes", ``No", ``Up",``Down", ``Left", ``Right", ``On", ``Off", ``Stop", ``Go", along with ``silence" (\textit{i.e.} no word spoken) and ``unknown" word, representing the remaining $20$ keywords from the dataset. The training set contains a total of $56,196$ clips with $32,550$ ($57 \%$) samples from the ``unknown" class and around $1800$ samples ($3.3 \%$) from each of the remaining classes, hence the dataset is naturally unbalanced. Also, we used the Common Voice Italian (CV Italian) dataset (version 6.1) containing a total of $84$K utterances ($132$ hours) which were recorded by more than $10$K Italian-speaking participants. The train set consists of $748$ speakers ($89$ hours of speech), while both valid and test sets contain around $22$ hours of speech from $1219$ and $3404$ speakers respectively. 

\textbf{Model Architectures.} For CIFAR10 and ImageNet we make use of ResNet-18 \citep{resnet}. For FEMNIST, we choose a much shallower CNN as proposed by~\citep{caldas2018leaf}. These architectures are kept the same for both centralized and FL experiments. These models are trained with SGD but only the centralized setting makes use of momentum. For the sake of completeness, we choose to use different deep learning model for the Speech Commands dataset. We employ $4$ layers of LSTM each with $256$ nodes. The models are trained using \textit{Adam} optimization. Also, the hyper-parameters, such as learning rates, are set to be the same as centralized learning without further tuning. For ASR task on CV Italian dataset, the experiments are based on a encoder-decoder model trained with the joint connectionist temporal classification (CTC)-attention objective \citep{kim2017joint}. A typical ASR model includes three modules: the encoder, the decoder and the attention mechanism. The encoder has the following architecture: CNN — LSTM — DNN, and the decoder is a single hidden layer GRU. Models are jointly trained with CTC and cross entropy (CE) loss. Note that the federated training for ASR task starts from a pre-trained initialized model since all the existing FL aggregation methods fail to converge without pre-training \citep{gao2022end, dimitriadis2020federated}.
%\hl{to update with new datasets and models. Mention that for COmmonVoice we rely on a pre-trained model. This is done because as seen in publications [A,B,C] training an ASR FL model from scratch does not work.}

\textbf{Data partition methodology.} As mentioned in Section \ref{secbackground}, FL settings can usually be classified as \textit{cross-silo} or \textit{cross-devices}. In \textit{cross-silo} settings, data distribution in each client will be the same as the global data distribution, hence training energy should be very close to centralized training with additional communication cost. In this work, we focus the experiments on \textit{cross-device} settings, and the IID partition provides the best-case scenarios and the baselines for comparison between centralized and FL settings.

%We simulate the difference in data distribution in these two scenarios by using two different partition schemes: a uniform partition (IID) where each client has approximately the same proportion of each class from the original dataset, and a heterogeneous partition (non-IID) for which each client has an unbalanced and different proportion of each class.
We simulate different level of non-IID data distribution following the latent Dirichlet allocation (LDA) partition method~\citep{reddi2020adaptive, yurochkin2019bayesian,hsu2019measuring} ensuring that each client gets allocated the same number of training samples. Each sample is drawn independently with class labels following a categorical distribution over $N$ classes parameterized with a vector \textbf{q} ($q_i \geq 0$, $i \in [1,m]$ and $\sum q_i = 1$ for a total of $m$ classes from the dataset). Thus, to simulate the partition, we draw $\textbf{q} \sim Dir(\alpha\textbf{p})$ from a Dirichlet distribution, where \textbf{p} stands for the prior distribution of the dataset, and $\alpha$ stands for the concentration which controls the level of heterogeneity of the partition. As $\alpha \to \infty$, the partition becomes more uniform (IID), and as $\alpha \to 0$, the partition becomes more heterogeneous. As the dataset is balanced across classes for both CIFAR10 and ImageNet, the prior distribution $\textbf{p}$ is uniform. For ImageNet, we chose $\alpha = 1000$ for the IID dataset partition and $\alpha = 0.5$ for non-IID following \citet{ yurochkin2019bayesian} and \citet{hsu2019measuring}. For CIFAR10, we choose $\alpha = 0.1$ following the same protocol as \citet{reddi2020adaptive}. As for Speech Commands, in light of the unbalanced nature of the dataset, we propose to change the prior of LDA from uniform distribution to multinomial distribution. Hence the LDA can be summarized as:
\begin{align}
    \label{eqprior}
    \textbf{p} &=\left(\frac{N_1}{N}, \frac{N_2}{N},...,\frac{N_m}{N}\right) \\
    \textbf{q} &\sim Dir(\alpha \textbf{p}),
\end{align}

where $N_i$ stands for the number of data from class $i$, $N$ stands for total number of data in the dataset. According to \citet{yurochkin2019bayesian,hsu2019measuring}, $\alpha$ is commonly set as $0.5$ for a non-IID partition of balanced dataset. Given the aforementioned unbalanced nature of the dataset, we propose to match the variance of $10$ keywords classes with multinomial prior to the variance of $10$ keywords classes with a uniform prior by changing $\alpha$ to $1.0$. %and then apply the prior distribution as described in Eq \ref{eq:prior}.

In practice, a non-IID dataset can mean both class-imbalance and feature-imbalance among clients. Other latent factors can change such as the user accent or voice timbre in speech recognition or different calligraphy styles in hand-written text. Therefore, we also include two naturally partitioned datasets FEMNIST and CV Italian to capture the feature imbalanced datasets.

For CV Italian, we first pre-train the model on half of the data samples in a centralized fashion. We do this by partitioning the original dataset into a small subset of speakers ($99$) for centralized training and a larger subset of speakers ($649$) for the FL experiment. Then, we simulate a scenario of single speaker using their individual devices by naturally dividing the training sets based on users ID into $649$ partitions. We followed the paritioning methodology in~\citet{caldas2018leaf} to extract the FEMNIST dataset from EMNIST following a natural partitioning by writer id.% FEMNSIT is also partitioned naturally by writers of digits and characters.

\textbf{Client pool.} Following \citet{reddi2020adaptive}, we consider a pool of $500$ client for CIFAR10 with $10$ active clients training concurrently per round.%Then, to investigate the impact of a different number of available clients, we conducted both ImageNet and SpeechCommands tasks with a set of $100$ clients, with $10$ sampled clients each round.
We split ImageNet and SpeechCommands into 100 clients and randomly select 10 clients per round. As for FEMNIST and CV Italian, there are $3597$ and $649$ natural clients respectively, and we select $35$ and $10$ clients in each communication round.

\textbf{FL strategy.} To better reflect realistic FL scenarios, we propose to investigate the energy consumption with the common FedAVG strategy \citep{pmlr-v54-mcmahan17a}, and the more complex FedADAM strategy \citep{reddi2020adaptive}. For CIFAR10, we follow the experimental protocol proposed in \citet{reddi2020adaptive} considering the suggested best values for $\eta$, $\eta_l$, and $\tau$ in almost every experiment except for FedAVG, where we had to lower the value of $\eta_l$ to $10^{-3/2}$ to allow training. All other experiments used a server learning rate $\eta = 0.1$ and $\tau= 0.001$.

\textbf{Local epoch (LE).} We also propose to vary the number of local epochs done on each client to better highlight the contribution of the local computations to the total emissions. To be consistent, we choose to do $1$ and $5$ local epochs across all tasks except ASR task (insisting with $5$ local epochs to obtain acceptable performance).

\textbf{Target accuracies.} To make fair comparisons between different setups, we set the target accuracies for each tasks and report the respective carbon emission. This is a common procedure when evaluating FL workloads. We set the target accuracies for CIFAR10, FEMNIST and ImageNet to be $70\%$, $80\%$ and $50\%$ top-1 accuracy respectively. For Speech Commands, the threshold is set to $70\%$, and for CV Italian, the target is set to be $25\%$ of Word Error Rate (WER). 
%\hl{As our primary objective is to estimate the carbon footprint, we leave reaching state-of-the-art performance as a secondary objective.}

%\subsection{Carbon footprint estimation}\label{sec:estimation}

\subsection{Experimental Results}
This section presents the experimental results. Power consumption and training times obtained for all FL and centralized setups are reported in Table \ref{tabpower}.  Table \ref{tabpower} also shows the power measurement and energy consumption for each setups. Both power usage and training time per epoch reflect the mean value for each training tasks. The total energy is calculated as the energy per device multiplied by the number of selected clients per communication round for FL. In the centralized scenario it is equal to the energy per device. The numbers of communication rounds required by each setup to reach their target accuracies are summarized in Table \ref{tabroundsPerStrategy}. Table \ref{tabcarbonresults} shows the carbon emission for each training task in every experimental setups, calculated by adding the energy consumption for communication and convert the energy consumption to carbon emission by multiplying the country-specific conversion factor as explained in Eq (\ref{eqcarboneq}) and Eq (\ref{eqcarboneq2}).

% THIS IS TABLE 1
\begin{table*}[ht]
    \centering
    \tablefontsize
    \scalebox{0.95}{
    \begin{tabular}{ p{1.3cm}ccccccccc}
    %\begin{tabular}{ p{1.3cm}p{1.0cm}p{1.2cm}p{1.2cm}p{1cm}p{1.2cm}c{1cm}c{1.1cm}c{1.6cm}c{1.6cm}}
        & & & & & & \multicolumn{4}{c}{\bf{Costs to Reach Threshold Accuracy}} \\
        \cmidrule{7-10}\\
        \multirow{2}{*}{\textbf{Dataset}} & \multirow{2}{*}{\parbox{0.8cm}{\centering\textbf{Training \\ Strategy}}} & \multirow{2}{*}{\parbox{0.8cm}{\centering\textbf{HW}}} & \multirow{2}{*}{\parbox{0.8cm}{\centering\textbf{Power \\Usage\\(W)}}} &  \multirow{2}{*}{\parbox{1.0cm}{\centering\textbf{Local\\Epochs}}} & \multirow{2}{*}{\parbox{1.0cm}{\centering \textbf{Time per Epoch(s)}}} &  \multirow{2}{*}{\centering\parbox{0.8cm}{\centering\textbf{Num.\\Rounds}}} & \multirow{2}{*}{\parbox{0.8cm}{\centering\textbf{Time(s)}}} & \textbf{Energy}  & \textbf{Total} \\
                         &                   &                   &                      &                       &                   &                 &                    & \textbf{per device} & \textbf{Energy} \\
                         &                   &                   &                      &                       &                   &                 &                    & \textbf{(Wh)} & \textbf{(Wh)} \\                 
        \midrule
        \multirow{6}{*}{CIFAR10} & Centralized & V100 & 160+42 & 1 & 24 & 2 & 48 & 2.7 & \textbf{2.7} \\
                                %  &                              & K80  & 240 + ? &                    & 42 &                    & 84 & 7.0 & 7.0 \\
        \cmidrule{2-10}
                                 & \multirow{2}{*}{FedAVG}      & TX2  & 4.7 & \multirow{2}{*}{5} & 0.8 & \multirow{2}{*}{580} & 2320 & 3.03 & 30.3 \\
                                 &                              & NX   & 6.3 &                    & 0.6 &                      & 1740 &  3.05 & 30.5 \\
        \cmidrule{2-10}
                                 & \multirow{2}{*}{FedAdam}     & TX2  & 4.7 & \multirow{2}{*}{1} & 0.8 & \multirow{2}{*}{1800} & 1440 & 1.88 & 18.8 \\
                                 &                              & NX   & 6.3 &                    & 0.6 &                      & 1080 & 1.89& 18.9 \\
        \midrule
        \multirow{6}{*}{ImageNet}&      Centralized             & V100 & 220+84 &      1             & 1,440 & 8 & 11,520 & 973 & \textbf{971} \\
        \cmidrule{2-10}
                                 & \multirow{2}{*}{FedAVG}      & TX2  & 6.5 & \multirow{2}{*}{1} & 474 & \multirow{2}{*}{339} & 160,686 & 290 & 2,901 \\
                                 &                              & NX   & 9.7 &                    & 273 &                      & 92,547 & 249 & 2,494 \\
        \cmidrule{2-10}
                                 & \multirow{2}{*}{FedAdam}     & TX2  & 6.5 & \multirow{2}{*}{1} & 474 & \multirow{2}{*}{590} & 279,660 & 504 & 5,049 \\
                                 &                              & NX   & 9.7  &                    & 273 &                      & 161,070 & 434 & 4,340 \\
        \midrule
        \multirow{5}{*}{FEMNIST}&      Centralized             & V100 & 96+20 &      1             & 19 & 1 & 19 & 0.6 & \textbf{0.6} \\
        \cmidrule{2-10}
                                 & \multirow{2}{*}{FedAVG}      & TX2  & 2.4 & \multirow{2}{*}{1} & 0.24 & \multirow{2}{*}{205} & 29 & 0.03 & 1.1 \\
                                 &                              & NX   & 2.7 &                    & 0.15 &                      & 18 & 0.02 & 0.8 \\
                                 \cmidrule{2-10}
                                 & \multirow{2}{*}{FedADAM}      & TX2  & 2.4 & \multirow{2}{*}{1} & 0.24 & \multirow{2}{*}{60} & 14 & 0.01 & 0.3 \\
                                 &                              & NX   & 2.7 &                    & 0.15 &                      & 9 & 0.007 & 0.2 \\
        \midrule
        \multirow{6}{*}{\parbox{1.4cm}{Speech \\ Commands}} & Centralized & V100 & 68+56 & 1 & 52 & 6 & 312 &  10.7 &  10.7 \\
                                %  &                              & K80  & 76 + ? &                    & 660 &                         & 5,280 & 440 & 440 \\
        \cmidrule{2-10}
                                 & \multirow{2}{*}{FedAVG}      & TX2  & 5.7 & \multirow{2}{*}{5} & 1.6 & \multirow{2}{*}{140} & 1,120 & 1.8 & 17.7 \\
                                 &                              & NX   & 7.9 &                    & 0.9 &                      & 630 & 1.4 & 13.8 \\
        \cmidrule{2-10}
                                 & \multirow{2}{*}{FedAdam}     & TX2  & 5.7 & \multirow{2}{*}{1} & 1.6 & \multirow{2}{*}{193} & 309 & 0.5 & 4.9 \\
                                 &                              & NX   & 7.9 &                    & 0.9 &                      & 174 & 0.4 & \textbf{3.8} \\
        \midrule
        \multirow{3}{*}{CV Italian}&      Centralized             & V100 & 170 + 48 &      1             & 509 & 10 & 5090 &  308.2 & \textbf{308} \\
        \cmidrule{2-10}
                                 & \multirow{2}{*}{FedAVG}      & TX2  & 6.7 & \multirow{2}{*}{5} & 76  & \multirow{2}{*}{50} & 19,000  & 35.4 & 354 \\
                                 &                              & NX   & 9.8 &                    & 48  &                     & 12,000 & 32.7 & 327 \\
        \bottomrule
    \end{tabular}
    }
    % \captionsetup{font=small,labelfont=bf}
    \caption{Energy consumption of centralized training using GPUs against FL settings where each client trains on a small dataset partition using low-power GPU-enabled edge devices. For FL rows, each strategy reports the lowest total energy among $1$ and $5$ local epochs for non-IID partitions. For centralized setting, one ``Local Epoch'' is one standard epoch using the entire dataset and, ``Power Usage'' is reported as GPU+CPU. The ``Time'' column reports the total training time required, which is calculated by multiplying the ``Time per Epoch'' and the ``Number of Rounds''. For FL rows, the ``Total Energy'' is obtained by multiplying the ``Energy per Device'' by the number of clients participating in each round. Despite edge devices consuming an order of magnitude less power, the total energy required for FL is often greater (but of the same order of magnitude) than centralized training. For Speech Commands, a very lightweight workload, FL can reach the target accuracy while requiring little energy. For datasets with variable amount of data per client (e.g. FEMNIST, CV Italian), we report the time taken to train a client that contains the average data samples observed in the whole dataset.}
    \label{tabpower}
\end{table*}

% THIS IS TABLE 2
\begin{table}[ht!]
    \centering
    \tablefontsize
    \scalebox{0.8}{
    \begin{tabular}{lcccc }
        \toprule
        \multirow{2}{*}{\textbf{Dataset}} & \textbf{Training} & \textbf{Local} & \multicolumn{2}{c}{\textbf{Partition}} \\
                & \textbf{Strategy} & \textbf{Epochs} &  \textbf{IID}  & \textbf{non-IID} \\
        \midrule
        \multirow{4}{*}{CIFAR10} & \multirow{2}{*}{FedAVG} & 1 & 480 & $>$2000 \\
                                 &                         & 5 & 180  & 580 \\
                                 \cmidrule{2-5}
                                 & \multirow{2}{*}{FedAdam}& 1 & 580 & 1800 \\
                                 &                         & 5 & 250  & 800 \\
        \midrule
        \multirow{4}{*}{ImageNet} & \multirow{2}{*}{FedAVG} & 1 & 232 & 339 \\
                                 &                         & 5 & 95  & 114 \\
                                 \cmidrule{2-5}
                                 & \multirow{2}{*}{FedAdam}& 1 & 550 & 590 \\
                                 &                         & 5 & 180  & 200 \\
        \midrule
        \multirow{4}{*}{FEMNIST} & \multirow{2}{*}{FedAVG} & 1 & - & 205 \\
                                 &                         & 5 & -  & 120 \\
                                 \cmidrule{2-5}
                                 & \multirow{2}{*}{FedAdam}& 1 & - & 60 \\
                                 &                         & 5 & -  & 40\\
        \midrule
        \multirow{4}{*}{\parbox{1.4cm}{Speech \\ Commands}} & \multirow{2}{*}{FedAVG} & 1 & $>$1000 & 770 \\
                                          &                         & 5 & 119  & 140 \\
                                          \cmidrule{2-5}
                                          & \multirow{2}{*}{FedAdam}& 1 & 140 & 193 \\
                                          &                         & 5 & 53  & 66 \\
        \midrule
        CV Italian & FedAVG & 5 & - & 50 \\
        \bottomrule
    \end{tabular}
    }
    \captionsetup{font=small,labelfont=bf}
    \caption{Number of FL rounds needed for each dataset-strategy pair to reach the target accuracy when data is partitioned in IID and non-IID fashion. Note that there is only non-IID partition for FEMNIST and CV Italian, as both datasets are naturally partitioned. We observe that increasing the number of local epochs always results in fewer FL rounds to reach convergence. However, this does not guarantee a smaller overall energy consumption.}
    \label{tabroundsPerStrategy}
\end{table}

As shown in Table \ref{tabpower}, it is worth noting that centralized training (V100) took solely $2$ epochs to achieve the target accuracy for CIFAR10, and $8$ epochs for ImageNet, $1$ for FEMNIST and $10$ for CV Italian. This translates to $48$ seconds for CIFAR10, $3.2$ hours for ImageNet, $19$ seconds for FEMNIST and $1.4$ hours for CV Italian.

Table \ref{tabroundsPerStrategy} reports the numbers of communication rounds required by each setup to reach their target accuracies. We can see that standard FedAVG failed to converge within the allotted $2000$ rounds in the non-IID setting when using only $1$ local epoch for CIFAR10, while the more sophisticated FedADAM strategy was able to reach the target. For Speech Commands experiments, it is interesting that FedAVG needs even more rounds for IID than non-IID if we only do one local epoch, which might be due to the dataset being naturally unbalanced. Similar as FEMNIST, CV Italian is a naturally partitioned dataset, so there only exist non-IID results. As settings with only $1$ local epoch does not converge, we only show settings with $5$ local epochs in the tables.

From Table \ref{tabcarbonresults} we can see that for image classification task (CIFAR10, ImageNet and FEMNIST) we observe the centralized settings generally consume less energy compared to their FL counterparts. The difference is the biggest when we compare CIFAR10 non-IID with $1$ local epoch settings with centralized training. In this comparison, FL emits more than $10$ times more carbon than centralized training. The difference is smaller when we perform $5$ local epochs in FL. However, for ImageNet, the outcome is the other way around. FL with $5$ local epochs emits more carbon compared to $1$ local epoch settings. The difference between FL and centralized training for ImageNet is smaller than CIFAR10.  For FEMNIST, as the dataset is naturally partitioned, there are only non-IID results. Similar as CIFAR10, $1$ local epoch settings emits more carbon compared with $5$ local epochs settings, and they both more emits higher carbon compared to centralized training. More surprisingly is the slower convergence rate of FedADAM for CIFAR10 and ImageNet to reach the specified target accuracy. However, FedADAM often performed better in the longer term resulting in higher final accuracies. For Speech Commands experiments, \ref{tabcarbonresults} also highlights the setups when FL emits less carbon compared to centralized training, which happens in France when FL performs $5$ local epochs. For the CV Italian experiments, it is worth noticing that all FL settings emits less carbon when compared with centralized training in the data centers with the averaged PUE ratio of $1.67$. It is even less than centralized training in the data centers with PUE ratio of $1.55$ in France.

%Table \ref{tab:speechtime} reports the experimental results of centralized learning for the Speech Commands dataset. Like the previous experiments, we measure the FL training time which is demonstrated in Table \ref{tab:speechpower}. Also, LSTM models can reach to maximum of $87 \%$ testing accuracy and we recorded the number of communication rounds required to reach the $70 \%$ threshold, which is shown in Table \ref{tab:speechround}, and the total carbon emission is reported in Table \ref{tab:carbonresults} including the WAN emission. 

\begin{table*}[!h]
\centering
\scalebox{0.7}{
%\begin{tabular}{ p{1.75cm}|p{1.4cm}| p{1.4cm}|p{0.7cm}p{0.7cm}| p{0.7cm}p{0.7cm}|p{0.7cm}p{0.8cm}| p{0.7cm}p{0.6cm}|p{0.65cm}p{0.7cm}|p{0.7cm}p{0.7cm}p{0.7cm}p{0.7cm}}
\begin{tabular}{ c|ccc|cccc|cccc|cccc}
\toprule
% CIFAR10
\textbf{CIFAR10} &  \multicolumn{3}{c|}{\centering\textbf{Centr.}} & \multicolumn{4}{c|}{\textbf{IID 5LE}} & \multicolumn{4}{ c| }{\textbf{non-IID 1LE}} & \multicolumn{4}{ c }{\textbf{non-IID 5LE}}\\
 \textbf{Country/} & \multicolumn{3}{ c| }{\textbf{PUE}} & \multicolumn{2}{ c }{\textbf{FedAVG}} & \multicolumn{2}{ c| }{\textbf{FedADAM}}& \multicolumn{2}{ c }{\textbf{FedAVG}} &\multicolumn{2}{ c| }{\textbf{FedADAM}}& \multicolumn{2}{ c }{\textbf{FedAVG}} &\multicolumn{2}{ c }{\textbf{FedADAM}}\\ 

 \textbf{CO$_2$e(g)} & 1.67 & 1.55 & 1.11 & TX2 & NX & TX2 & NX & TX2 & NX & TX2 & NX & TX2 & NX & TX2 & NX \\
\midrule
Australia & 3.0 & 2.7&  \textbf{2.0}  & 70.6 & 78.1 & 98.1 & 108.5  & $>$730 & $>$813 & 656.7 & 731.6 & 227.5 & 251.7 & 313.8 & 347.2\\
UK & 1.3& 1.2& \textbf{0.8}& 29.4 &32.5&40.8& 45&$>$303&$>$337.7&272.8&303.9&94.7&104.8&130.6&144.5 \\
France & 0.2&0.2 &\textbf{0.2}& 2.1 & 2.3 & 3.0 &3.2&$>$19&$>$21& 17.4 &19.3& 6.9 & 7.5 & 9.5 & 10.4\\
\bottomrule
\end{tabular}
}

\vspace{0.25cm}

\scalebox{0.7}{
\begin{tabular}{ c|ccc|cccc|cccc|cccc}
\toprule
\textbf{ImageNet} & \multicolumn{3}{c|}{\centering\textbf{Centr.}} & \multicolumn{4}{c|}{\textbf{IID 5LE}} & \multicolumn{4}{ c| }{\textbf{non-IID 1LE}} & \multicolumn{4}{ c }{\textbf{non-IID 5LE}}\\
 \textbf{Country/} & \multicolumn{3}{ c| }{\textbf{PUE}} & \multicolumn{2}{ c }{\textbf{FedAVG}} & \multicolumn{2}{ c| }{\textbf{FedADAM}}& \multicolumn{2}{ c }{\textbf{FedAVG}} &\multicolumn{2}{ c| }{\textbf{FedADAM}}& \multicolumn{2}{ c }{\textbf{FedAVG}} &\multicolumn{2}{ c }{\textbf{FedADAM}}\\ 

 \textbf{CO$_2$e(g)} & 1.67&1.55 & 1.11 & TX2 & NX & TX2 & NX & TX2 & NX & TX2 & NX & TX2 & NX & TX2 & NX \\
\midrule
Australia & 1066 & 989 & \textbf{708}&2701& 2330 & 5117 & 4415 & 2025 & 1771 & 3524 & 3083 & 3241 & 2796 & 5686 & 4905 \\
UK & 457&424&\textbf{303}&1156&998&2191 & 1890 &866 &757 &1507 &1317 &1388 &1197 &2435 &2100\\
France &88&81& \textbf{59}&  220 & 190 & 418 & 359 & 160 & 138 & 278 & 240 & 265 & 228 & 464 & 399 \\
\bottomrule
\end{tabular}
}

\vspace{0.25cm}

\scalebox{0.7}{
\begin{tabular}{ c|ccc|cccc|cccc}
%{ p{1.75cm}|p{2cm}|p{2cm}|p{1.8cm}p{1.8cm}| p{1.8cm}p{1.8cm}| p{1.8cm}p{1.8cm}}
\toprule
\textbf{FEMNIST}& \multicolumn{3}{ c| }{\textbf{Centr.}}& \multicolumn{4}{ c|}{\textbf{non-IID 1LE}}&\multicolumn{4}{ c }{\textbf{non-IID 5LE}}\\

 \textbf{Country/}  &\multicolumn{3}{ c| }{\textbf{PUE}}& \multicolumn{2}{ c }{\textbf{FedAVG}} & \multicolumn{2}{ c| }{\textbf{FedADAM}}& \multicolumn{2}{ c }{\textbf{FedAVG}}& \multicolumn{2}{ c }{\textbf{FedADAM}} \\ 

 \textbf{CO$_2$e(g)} & 1.67 &1.55 & 1.11 & TX2 & NX& TX2 & NX& TX2 & NX& TX2 & NX \\
\midrule
Australia & 0.7 & 0.6 &\textbf{0.4}& 140.9 & 156.9   & 41.2 & 45.9 & 84.2 & 93.1 & 28.1 &30.1\\
UK & 0.3 & 0.3 &\textbf{0.2} & 58.5 & 65.1 &17.1 & 19.1 & 35.0 & 38.7 &11.7 & 12.9\\
France & 0.1 & 0.1& \textbf{0.03}  & 3.6 & 4.0 & 1.1  & 1.2 & 2.3 & 2.5 & 0.8 &0.8\\
\bottomrule
\end{tabular}
}

\vspace{0.25cm}

\scalebox{0.65}{
\begin{tabular}{ c|ccc|cccc|cccc|cccc}
%{ p{2.5cm}|p{1.3cm}| p{1.3cm}|p{0.7cm}p{0.7cm}| p{0.7cm}p{0.7cm}|p{0.7cm}p{0.7cm}| p{0.7cm}p{0.7cm}|p{0.7cm}p{0.7cm}|p{0.7cm}p{0.7cm}p{0.7cm}p{0.7cm}}
\toprule
\textbf{SpeechCmd}& \multicolumn{3}{ c| }{\textbf{Centr.}}& \multicolumn{4}{ c| }{\textbf{IID 5LE}} &\multicolumn{4}{ c| }{\textbf{non-IID 1LE}} & \multicolumn{4}{ c }{\textbf{non-IID 5LE}}\\

 \textbf{Country/}  &\multicolumn{3}{ c| }{\textbf{PUE}}&\multicolumn{2}{ c }{\textbf{FedAVG}} &\multicolumn{2}{ c |}{\textbf{FedADAM}}& \multicolumn{2}{ c }{\textbf{FedAVG}} &\multicolumn{2}{ c| }{\textbf{FedADAM}}& \multicolumn{2}{ c }{\textbf{FedAVG}} &\multicolumn{2}{ c }{\textbf{FedADAM}}\\ 

 \textbf{CO$_2$e(g)} & 1.67 &1.55 & 1.11  & TX2 & NX & TX2 & NX & TX2 & NX & TX2 & NX & TX2 & NX & TX2 & NX \\
\midrule
Australia & 11.8 &10.9  & \textbf{7.8}  & 30.5 & 30.8 &13.6& 13.7& 146.4 & 159.1 & 36.7 & 39.9 & 35.9 & 36.2 & 16.9 & 17.1\\
UK & 5.0  & 4.7 & \textbf{3.4}  & 12.8 & 12.9 & 5.7 & 5.7 & 60.9 & 66.2 
& 15.3 & 16.6 & 15.1 & 15.1 & 7.1 & 7.1 \\
France & 1.0 &0.9 & 0.6  & 1.3 & 1.2 & 0.6 & \textbf{0.5} & 4.4 & 4.6 & 1.1 & 1.2 & 1.6 & 1.4 & 0.7 & 0.7\\
\bottomrule
\end{tabular}
}

\vspace{0.25cm}
\scalebox{0.7}{
\begin{tabular}{ c|ccc|cc}
%{ p{1.75cm}|p{2cm}|p{2cm}|p{1.8cm}p{1.8cm}| p{1.8cm}p{1.8cm}| p{1.8cm}p{1.8cm}}
\toprule
\textbf{CV Italian}& \multicolumn{3}{ c| }{\textbf{Centr.}}& \multicolumn{2}{ c }{\textbf{non-IID 5LE}}\\

 \textbf{Country/}  &\multicolumn{3}{ c| }{\textbf{PUE}} & \multicolumn{2}{ c }{\textbf{FedAVG}} \\ 

 \textbf{CO$_2$e(g)} & 1.67 &1.55 & 1.11 & TX2 & NX \\
\midrule
Australia & 337.7 &313.4  &224.4  & 330.3 & \textbf{324.0} \\
UK & 144.6& 134.2 & 96.1 &  140.2 & \textbf{137.3}\\
France & 27.8 &25.8& 18.5 & 21.6 & \textbf{20.4}\\
\bottomrule
\end{tabular}
}

\captionsetup{font=small,labelfont=bf}
\caption{CO$_2$e emissions (expressed in grams, i.e \textbf{lower is better}) for both centralized learning and FL when they reach the target accuracies, with different tasks and setups. The tables report results of both FedAvg and FedADAM in both IID and non-IID partitions. As non-IID is more realistic, we report both $1$ and $5$ local epochs experiment results for this setup only. Results in bold indicate lower carbon emissions overall. }
\label{tabcarbonresults}
\end{table*}

\section{Carbon Footprint of Federated Learning}\label{secanalysis}
\subsection{CO$_2$e Analysis}

So far we have considered the energy required to achieve a given accuracy on different tasks for various sets of hyper-parameters and optimizers. We now turn our attention to how this translates into carbon emissions.

The first thing to notice is that there are some settings with Speech Commands and CV Italian where FL emits slightly less carbon compared with centralized training. For Speech Commands, the model architecture is light-weighted, hence both training using TX2 and NX consumed much less energy, as shown in Table \ref{tabpower}. Since the model only has $5.3$ million parameters, communication did not consume much energy either. As for CV Italian, training energy for FL and centralized is about the same as shown in Table \ref{tabpower}. Since the process only requires $50$ communication rounds to reach our target accuracy and because we need to take into account the PUE ratio for data centers, the overall carbon emission for FL,  in this specific scenario, can be lower than centralized training. Therefore, emissions from centralized and federated learning can be more comparable when using lightweight models, typically of cross-device setups. 

Due to the large difference between electricity-specific CO$_2$e emission factors among countries, the carbon footprint of both centralized training and FL can be highly dependent on the geolocation of hardware. Training in France always has the lowest CO$_2$e emissions given their use of nuclear energy with the lowest energy to CO$_2$e conversion rate. Geolocation also impacts the carbon footprint of training in FL via  communication speed. If the physical location has a slower Internet connection, the total time for communicating model parameters back and forth from the clients to the server will be longer, hence more energy is consumed.
%FL can be comparable or even greener than centralized training. As seen in Table \ref{tab:carbonresults}, training on CIFAR10 in France using FL can save $0.3$g CO$_2$e when compared to centralized training in China.  %$0.3$g to $4.4$g CO$_2$e if compared to centralized training in China. 
%For larger datasets, such as ImageNet, this assertion becomes true for any FL setup in France to any centralized setup located in China and the US. Finally, on the Speech Commands dataset, FL is more efficient than centralized training if using FedADAM with 5LE in any given countries. 

%Also, CO$_2$e emissions induced by FL highly depend on other factors as well, such as DL task and model architecture. Indeed, we show that a basic FL setup relying on FedAVG clearly emits more carbon for image-based tasks using ResNet-18 compared to modern GPUs and centralized training, while it also emits less carbon in some setups with the Speech Commands dataset using a stacked LSTM model. As a matter of fact, the lower CO$_2$e estimates obtained with FL on Speech Commands are of great interest. Indeed, in practice, it is most likely that deployed FL tasks (i.e on device) will be lightweight and certainly aiming at lowering the number of communications. Hence, this type of realistic FL use-cases could \textit{benefit} to our planet.

Hardware efficiency is also a critical factor when estimating the total carbon footprint. As new AI applications for consumers are created every day, it is realistic to assume that novel versions of chips like Tegra TX2 will soon be embedded in numerous devices, including smartphones, tablets, and others. However, such specialized hardware is certainly not an exact estimate of what is currently being used for FL. Therefore, to facilitate carbon impact estimations of large-scale FL deployment, the industry must increase its transparency with respect to its devices' distribution over the market. As we can see from the results, even though NX requires less training time compared to TX2, it also consumes more power both during training and in an idle state. This leads to a trade-off between high-power hardware and actually training consumption. For example, training FEMNIST with 1 local epoch with FedAVG in TX2 emits more carbon compared to NX, but it emits less carbon compared to NX when we switch to FedADAM. %Of course, both FL and centralized learning benefit from more efficient hardware. 

As explained in our estimation methodology, FL will always have an advantage in the respect that FL does not require cooling as opposed to centralized learning in the data centers.  In fact, even though GPUs or even TPUs are getting more efficient in terms of computational power delivered by the amount of energy consumed, the need for strong and energy-consuming cooling remains; thus, the FL can always benefit more from the hardware advancement. On the other hand, FL always has a drawback of communication when the model parameters are communicated between clients and the central server. 
%advantage only grows. Unlike centralized learning, FL always benefits from a net decrease in CO$_2$e emission each time the hardware is improved. 

\begin{figure*}[ht]
\centering
\begin{tabular}{p{4.8cm}p{4.6cm}p{4.5cm}}

\begin{tikzpicture}[scale = 0.6]
% CIFAR10
\begin{axis}[ymode=log,
    axis lines = left,
    xlabel = Number of centralized epoch,
    ylabel = {log(CO$_2$e emitted in g) },
    legend style={at={(1.8,1.4)},anchor=north,legend columns=4},
    title= (a) CIFAR10,
    grid=major,
]
%FL 5LE
\addplot [
    domain=0:50, 
    samples=200, 
    color=orange,
    grid=major,
    line width=1.5pt,
]
{1.63*x };
\addlegendentry{FL (5 local epochs)}

%FL 1LE
\addplot [
    domain=0:50, 
    samples=2000, 
    color=red,
    grid=major,
    line width=1.5pt,
]
{7.58*x };
\addlegendentry{FL (1 local epochs)}

%V100
\addplot [
    domain=0:50, 
    samples=100, 
    color=blue,
    grid = major,
    line width=1.5pt,
    ]
    {0.63*x};
\addlegendentry{NVIDIA Tesla V100}
    
\addplot[color = green, dashed, line width=1.5pt, mark=*] coordinates {
    (2,1.3)
    (0.77369,1.3)
    (0.16678,1.3)
    (0,1.3)
    };
    \addlegendentry{Breakeven line}
    
%\addplot[color = brown, dashed, line width=1.5pt,mark=*] coordinates {
%    
%    };
%    \addlegendentry{Breakeven line}

\end{axis}
%\node[anchor=north] at={(1,1)} { (a) CIFAR10};
\end{tikzpicture}

&
%ImageNet
\begin{tikzpicture}[scale = 0.6]
\begin{axis}[ymode=log,
    axis lines = left,
    xlabel = Number of centralized epoch,
    title= (b) ImageNet,
    grid=major,
]
%FL 5LE
\addplot [
    domain=0:50, 
    samples=100, 
    color=orange,
    grid=major,
    line width=1.5pt,
]
{24.34*x };
%\addlegendentry{FL (5 local epochs)}

%FL 1LE
\addplot [
    domain=0:50, 
    samples=100, 
    color=red,
    grid = major,
    line width=1.5pt,
    ]
    {25.52*x};
%\addlegendentry{NVIDIA Tesla V100}

%V100
\addplot [
    domain=0:50, 
    samples=100, 
    color=blue,
    grid = major,
    line width=1.5pt,
    ]
    {57.06*x};
%\addlegendentry{NVIDIA Tesla V100}

\addplot[color = green, dashed, line width=1.5pt,mark=*] coordinates {
    (18.75,456.5)
    (8,456.5)
    (0,456.5)
    };
    %\addlegendentry{FL (5 local epochs) reaches target accuracy}

\end{axis}
\end{tikzpicture}

&

% Speech Commands
\begin{tikzpicture}[scale = 0.6]
\begin{axis}[ymode=log,
    axis lines = left,
    xlabel = Number of centralized epoch,
    title= (c) Speech Commands,
    grid=major,
]
%FL 5LE
\addplot [
    domain=0:80, 
    samples=100, 
    color=orange,
    grid=major,
    line width=1.5pt,
]
{0.2152*x };
%\addlegendentry{FL (5 local epochs)}

%FL 1LE
\addplot [
    domain=0:80, 
    samples=100, 
    color=red,
    grid = major,
    line width=1.5pt,
    ]
    {0.7915*x};
%\addlegendentry{NVIDIA Tesla V100}

%V100
\addplot [
    domain=0:80, 
    samples=100, 
    color=blue,
    grid = major,
    line width=1.5pt,
    ]
    {0.84*x};
%\addlegendentry{NVIDIA Tesla V100}

\addplot[color = green, dashed, line width=1.5pt,mark=*] coordinates {
    (23.4,5.0)
    (0,5.0)
    %(113,10)
    (6.37,5.0)
    };
    %\addlegendentry{FL (5 local epochs) reaches target accuracy}

\end{axis}
\end{tikzpicture}

\end{tabular}
\captionsetup{font=small,labelfont=bf}
\vspace{-0.15cm}
\caption{Growth of CO$_2$e emissions (in log scale) with a $PUE=1.67$ for centralized learning and TX2 devices for FL in the UK (expressed in grams, i.e \textbf{lower is better}). Communication rounds in FL are converted to centralized epochs for a fair comparison, and CO$_2$e emissions are linearly dependent on the number of centralized epochs. The break-even line is chosen at the level when the centralized training reaches target accuracy. (a) For CIFAR10, 1 centralized epoch is equivalent to 50 communication rounds for 1 Local Epoch (LE) and 10 for 5 LE, as there are a total of $500$ clients. The green line shows that with an equal amount of emissions between FL and V100, FL would train for $7.5$ rounds with 5LE and $8$ rounds with 1LE. (b) For ImageNet, due to the smaller size of the total client pool ($100$), 1 centralized epoch is equivalent to 2 communication rounds with 5LE and 10 rounds with 1LE. The green line shows that with an equal amount of emissions, FL can only train for $178$ rounds with 1LE and $38$ rounds with 5LE. (c) For Speech Commands, the total number of clients is also $100$. The green line shows that with an equal amount of emissions, FL can only train for $47$ rounds with 1LE and $64$ rounds with 5LE.}
\label{figresultsgrowth}
\end{figure*}

Furthermore, CO$_2$ emissions depend on the distribution of clients' datasets. Our results show that realistic training conditions for FL (\textit{i.e.} non-IID data) are largely responsible for longer training times, which in turn translates to a high level of CO$_2$e emissions. While it is well known that the simpler aggregation form of FL (\textit{e.g. }FedAVG) performs reasonably well on IID data, it definitely struggles with non-IID partitioned data in terms of accuracy \citep{fedprox, qian2020towards}. %we extended the analysis to the more effective FedADAM strategy.
Interestingly, more complex strategies such as FedADAM can enable a decrease of up to $75\%$ and $70\%$ of the emitted CO$_2$e on Speech Commands and FEMNIST, respectively, compared to FedAvg. It is worth pointing out that for CIFAR10, FEMNIST, and Speech Commands non-IID partitions running 1LE produces more carbon than 5LE regardless of the aggregation strategy or devices. This is because communication consumption plays a big role in the total energy consumption, and 5LE communication costs less than using 1LE as fewer communication rounds are required. %However, we also found the FedADAM makes it crucial to carefully tune hyper-parameters to outperform FedAVG, especially for more challenging datasets such as ImageNet.

Figure \ref{figresultsgrowth} shows the growth of carbon emission when the number of centralized epochs increases. We first see that for CIFAR10, FL with 1LE has the highest slope, while for the other two datasets, centralized learning has the highest slope. Normally centralized learning should exhibit stronger slopes as TDPs of centralized learning hardware are much higher than for FL. However, for CIFAR10, and due to the large model size, the communication consumption is much higher than the actual training consumption, resulting in a very steep slope suggesting that employing a complex model does not benefit FL. 

%While it is clear that FedADAM improves the convergence rate, hence reducing the overall emission, it is still clearly not as ``green" as centralized learning for image tasks.
In Figure \ref{figcarbon_vs_acc}, we compare equivalent carbon budgets on CIFAR for FL and V100s. The former would only be able to train for $7.5$ and $8$ rounds with $5$ and $1$ local epochs, respectively, resulting in degraded performances. The same goes for ImageNet. Indeed, FL would only train for $178$ rounds with $1$ LE and $38$ rounds with $5$ LE. On the other hand, in Speech Commands, FL did not outperform in the UK. Hence would only train for $64$ rounds with $1$ LE and $47$ with $5$ LE. However, we can see that the difference between the break-even rounds and actual rounds required, as shown in Table \ref{tabroundsPerStrategy}, is much smaller than other tasks. It is also interesting to note that, for ImageNet, the communication cost is negligible, hence both FL curves look very similar.

\begin{figure*}[ht]
\centering
\begin{tabular}{p{4.8cm}p{4.6cm}p{4.5cm}}

\begin{tikzpicture}[scale = 0.6]
% CIFAR10
\begin{axis}[
    axis lines = left,
    xlabel = Accuracy,
    ylabel = {CO$_2$e emitted in g },
    legend style={at={(1,1.4)},anchor=north,legend columns=4},
    title= (a) CIFAR10,
    grid=major,
]

%IID
\addplot[color = green, dashed,line width=1.5pt, mark=*] coordinates {
    (0,0)
    (0.4,6.5)
    (0.5,18)
    (0.60,26.1)
    (0.70,40.8)
    };
    \addlegendentry{IID}

%nonIID
\addplot[color = red, line width=1.5pt, mark=*] coordinates {
    (0,0)
    (0.40,35.9)
    (0.50,62.1)
    (0.60,78.4)
    (0.70,130.6)
    };
    \addlegendentry{non-IID}

\end{axis}
%\node[anchor=north] at={(1,1)} { (a) CIFAR10};
\end{tikzpicture}

&
%ImageNet
\begin{tikzpicture}[scale = 0.6]
\begin{axis}[
    axis lines = left,
    xlabel = Accuracy,
    title= (b) ImageNet,
    grid=major,
]

\addplot[color = green, dashed,line width=1.5pt,mark=*] coordinates {
    (0,0)
    (0.1,365)
    (0.2,609)
    (0.3,974)
    (0.4,1339)
    (0.5,2191)
    };
\addplot[color = red, line width=1.5pt,mark=*] coordinates {
    (0,0)
    (0.1,365)
    (0.2,609)
    (0.3,974)
    (0.4,1339)
    (0.5,2434)
    };

\end{axis}
\end{tikzpicture}

\end{tabular}
\captionsetup{font=small,labelfont=bf}
\vspace{-0.15cm}
\caption{Growth of CO$_2$e emissions in the UK using TX2 for CIFAR10 and ImageNet respects to accuracies. The reported CO$_2$e emissions is for FedADAM with $5$ local epochs.}
\label{figcarbon_vs_acc}
\end{figure*}

Furthermore, Fig. \ref{figcarbon_vs_acc} demonstrates the growth of carbon emission with respect to accuracies. Non-IID partitioning generally emits more carbon as it requires larger numbers of communication rounds. Fig. \ref{figcarbon_vs_acc} also shows that the marginal carbon emission for additional accuracy gains is increasing exponentially. However, it is interesting to notice that carbon emissions of IID and non-IID at the beginning of ImageNet training overlap, as they require a similar number of rounds to reach certain accuracies. 

Finally, it is also worth noticing that the percentage of CO$_2$e emission resulting from WAN changes across the dataset and FL setups. It highly depends on the size of the model, the size of the dataset in each client, and the energy consumed by clients during training. More precisely, communications accounted for up to $0.7\%$ (ImageNet with 5 local epochs) and $96\%$ (CIFAR10 with 1 local epoch) of the total emissions. With CIFAR10 tasks, communication actually emits much more CO$_2$e than training, while on the other hand, WAN plays a very small role for ImageNet.

\subsection{Road-map for FL}

FL is still a maturing framework with a lot to improve in a different aspect. We would like to highlight a few challenges and future research directions based on our analysis.

First, as carbon footprint largely depends on the physical location of hardware, either in terms of training or communication, carbon emission can be immensely reduced by selecting clients from greener locations or with faster internet connections. Obviously, there will be practical concerns in choosing clients in certain locations more often. For example, clients from greener locations might not have enough data samples for training or might represent a skewed data distribution. This, however, could lead to a demographic bias and  needs further investigation.  

Also, industrial statistics on the available fleet of devices are crucial to optimize the carbon emissions of FL. Indeed, in the real world, hardware efficiency can vary vastly from client to client. Similarly to the physical location, we would also like to choose clients with more efficient hardware and comparable computing capability and such a selection also induces potential biases.

As is the case in centralized training, hyper-parameter tuning is of great importance in reducing training times. % In our experiments, we set the hyper-parameters the same as centralized training in each task, 
In our experiments, we decided only to modify optimizer-related parameters (e.g. learning rate, momentum) to ensure a fair comparison and a sufficient level of performance. Further tuning can be done to facilitate the training convergence of FL. Nevertheless, with FL, hyper-parameter tuning becomes a more arduous task as it potentially involves hundreds of different models (i.e. local clients models), each making use of a small dataset that is likely to follow a very skewed distribution. In addition to client-side tuning, the aggregation strategy (e.g. FedADAM) might also offer further parameterization, therefore increasing the complexity of the tuning process. 
%Unfortunately, tuning hyper-parameters can quickly become time and energy consuming, thus resulting in even more carbon emissions.
Therefore, novel hyper-parameter tuning algorithms should carefully be designed to minimize carbon emission by jointly maximizing the accuracy and minimizing the released CO$_2$e. 

The number of local epochs is also an important hyper-parameter that can surely impact the overall carbon emission. As seen in Table \ref{tabcarbonresults}, $5$ local epochs settings often emit less carbon than $1$ local epoch settings for non-IID, apart from the ImageNet task. This is easily explained by the hidden communication cost. Indeed, a single local epoch implies more communication rounds and, therefore, energy to converge compared to five local epochs. Furthermore, the number of communication rounds required for five local epochs usually is less than five times the number of communication rounds required for one local epoch. In the context of ImageNet, things are completely different as the local training becomes much more energy-demanding. Therefore, simply finding the right number of local epochs also clearly appears as a critical point in reducing FL carbon emissions.

Finally, carbon emission also depends on aggregation strategies. With more advanced aggregation strategies, the number of communication rounds can be reduced, hence reducing the overall carbon emissions. 

In summary, we quantify the carbon footprint based on the training energy consumption and communication energy consumption, which depend on the physical locations of the hardware, hardware efficiency, training hyper-parameters, and FL strategies.  We found that the carbon footprint of FL is hard to assess compared to centralized training without context, due to the inherent complications in how FL is currently performed. The complications might include data heterogeneity, client geographic distribution, and system heterogeneity. We provide a comprehensive analysis in this section and highlight the challenges and future research directions toward a more carbon-friendly FL.

%Since the model is relatively small and training time per local epoch is very efficient as shown in Table \ref{tab:power}, FL performs a lot better than centralized learning in terms of carbon emission even when compared to the more efficient hardware V100. As in previous experiments, we can notice that more efficient strategy like FedADAM can reduce the carbon emission by more than $50\%$ regardless of specific FL setup. It is worth pointing out that for FedAVG, non-IID with $1$ local epoch setup produces more carbon than $5$ local epochs setup, while for FedADAM, it is the other way around. Also, with $1$ local epoch setup, more communication rounds is required hence more WAN energy is consumed. WAN emission can be as high as $50\%$ of the total emission (non-IID with $1$ local epoch setup).

\section{Conclusion}
% Centralized is playing an increasing role in this Climate change, 
Federated learning is an upcoming paradigm in the ML world that is often proposed as an alternative to an already carbon-emitting centralized training. A number of recent studies have begun to detail the environmental costs of their novel deep learning methods, sometimes even integrating CO$_2$e emissions as an objective to be minimized. Following this important trend, this article takes a first look into the carbon footprint of an increasingly deployed training strategy known as federated learning. In particular, this work introduces a generalized methodology to systematically compute the carbon footprint of many FL setups and conducts extensive experiments on real FL hardware under different settings, models, strategies, and tasks. We highlight the carbon footprint of FL from different perspectives and demonstrate that each element of FL can have an impact on the total CO$_2$e emission, including physical location, deep learning tasks, model architecture, FL aggregation strategies, and hardware efficiency. Finally, we hope to emphasize the importance of taking the carbon footprint into consideration for future research, and innovative research for both deep learning and FL can integrate the carbon footprint as a novel metric.

\acks{This work was supported by the UK's Engineering and Physical Sciences Research Council (EPSRC) with grants EP/M50659X/1 and EP/S001530/1 and the European Research Council via the REDIAL project.
}

% Manual newpage inserted to improve layout of sample file - not
% needed in general before appendices/bibliography.

%\newpage

\appendix

\bibliography{21-0445}

\end{document}